\documentclass[runningheads]{llncs}

\usepackage[utf8]{inputenc}
\usepackage{amsmath}
\usepackage{booktabs}
\usepackage{array}
\usepackage{multirow}
\usepackage{tabularx}
\usepackage{longtable}
\usepackage{adjustbox}
\usepackage{subcaption}
\usepackage{makecell}
\usepackage{diagbox}
\usepackage{rotating}
\usepackage{wrapfig}
\usepackage{lipsum}
\usepackage{changepage}
\usepackage{epigraph}
\usepackage{tcolorbox}
\usepackage{placeins}
\usepackage{xcolor}
\usepackage{listings}
\usepackage{soul}
\usepackage{hyphenat}
\usepackage{comment}
\usepackage{graphicx}
\usepackage{hyperref}

\DeclareUnicodeCharacter{00B1}{\ensuremath{\pm}}
\DeclareUnicodeCharacter{00B2}{\textsuperscript{2}}
\DeclareUnicodeCharacter{00D7}{\ensuremath{\times}}
\DeclareUnicodeCharacter{2191}{\ensuremath{\uparrow}}
\DeclareUnicodeCharacter{2192}{\ensuremath{\rightarrow}}
\DeclareUnicodeCharacter{2193}{\ensuremath{\downarrow}}
\DeclareUnicodeCharacter{2212}{-}

\lstdefinestyle{logstyle}{
  basicstyle=\ttfamily\scriptsize,
  backgroundcolor=\color{gray!10},
  frame=single,
  framerule=0pt,
  lineskip={-0.5pt},
  breaklines=true,
  showstringspaces=false,
  numbers=left,
  numberstyle=\tiny\color{gray},
  stepnumber=1
}

\begin{document}

\title{\textbf{\textit{FSKD}}: Monocular \textbf{\textit{F}}orest \textbf{\textit{S}}tructure Inference via LiDAR-to-RGBI \textbf{\textit{K}}nowledge \textbf{\textit{D}}istillation}
\titlerunning{FSKD}
\author{Taimur Khan\inst{1,2}\thanks{Corresponding author.} \and Hannes Feilhauer\inst{2} \and Muhammad Jazib Zafar\inst{3}}
\authorrunning{T. Khan et al.}
\institute{Helmholtz Centre for Environmental Research -- UFZ, Halle (Saale), Germany\\
\email{taimur.khan@ufz.de}
\and
Leipzig University, Leipzig, Germany
\and
Georg-August University of G\"ottingen, G\"ottingen, Germany}

\maketitle

  \begin{abstract}
 Very High Resolution (VHR) forest structure data at individual-tree scale is essential for carbon, biodiversity, and ecosystem monitoring. Still, airborne LiDAR remains costly and infrequent despite being the reference for forest structure metrics like Canopy Height Model (CHM), Plant Area Index (PAI), and Foliage Height Diversity (FHD). We propose \textit{FSKD}: a LiDAR-to-RGB-Infrared (RGBI) knowledge distillation (KD) framework in which a multi-modal teacher fuses RGBI imagery with LiDAR-derived planar metrics and vertical profiles via cross-attention, and an RGBI-only SegFormer student learns to reproduce these outputs. Trained on 384 km² of forests in Saxony, Germany (20 cm ground sampling distance (GSD)) and evaluated on eight geographically distinct test tiles, the student achieves state-of-the-art (SOTA) zero-shot CHM performance (MedAE 4.17 m, $R^2$=0.51, IoU 0.87), outperforming HRCHM/DAC baselines by 29--46\% in MAE (5.81 m vs. 8.14--10.84 m) with stronger correlation coefficients (0.713 vs. 0.166--0.652). Ablations show that multi-modal fusion improves performance by 10--26\% over RGBI-only training, and that asymmetric distillation with appropriate model capacity is critical. The method jointly predicts CHM, PAI, and FHD, a multi-metric capability not provided by current monocular CHM estimators, although PAI/FHD transfer remains region-dependent and benefits from local calibration. The framework also remains effective under temporal mismatch (winter LiDAR, summer RGBI), removing strict co-acquisition constraints and enabling scalable 20 cm operational monitoring for workflows such as Digital Twin Germany and national Digital Orthophoto programs.

\textbf{Code + Models:} TBA upon publication.

\textbf{Data:} TBA upon publication.
\end{abstract}

\section{Introduction}

\begin{figure}[!htb]
    \centering
    \includegraphics[width=1\textwidth]{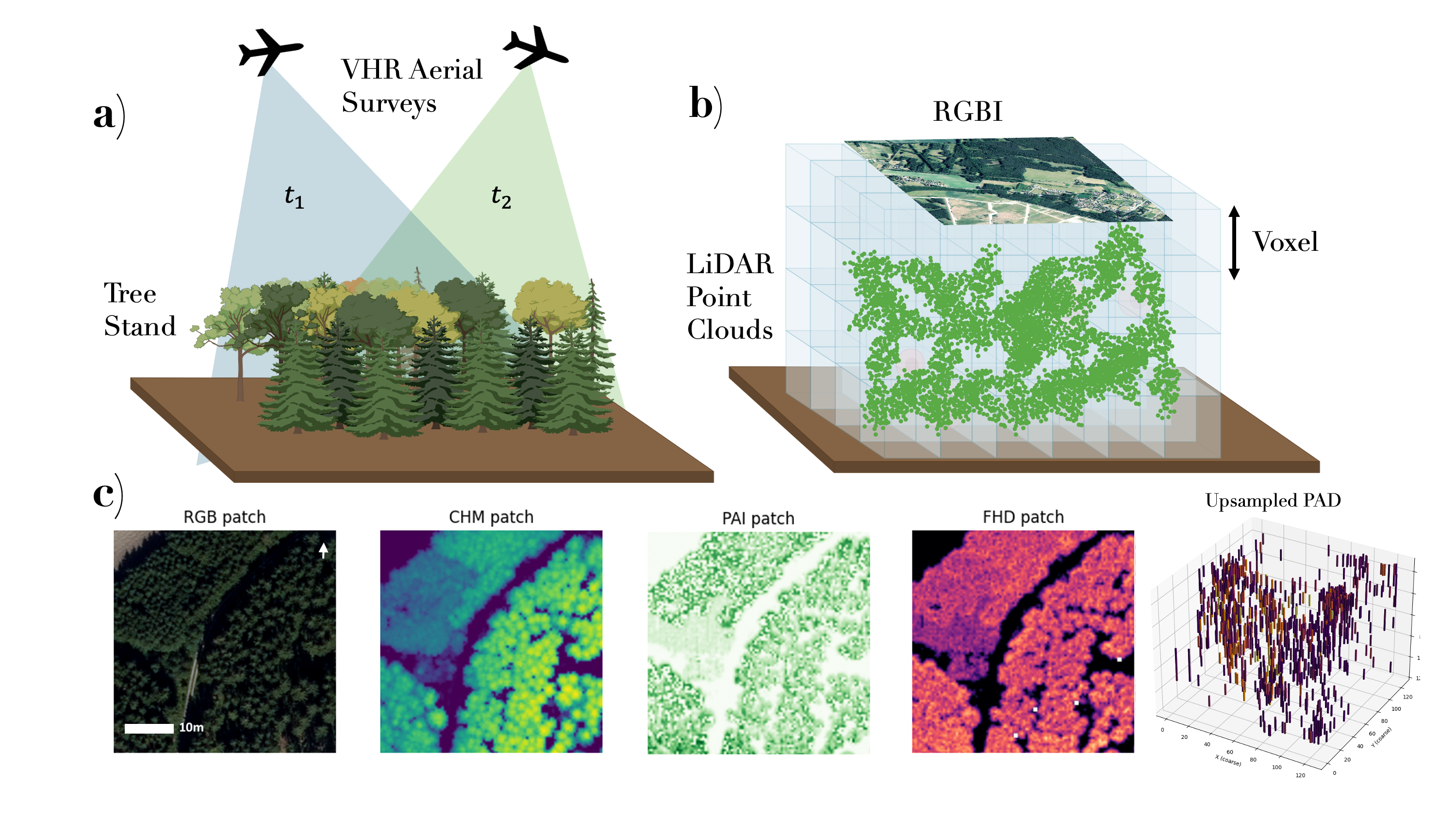}
    \caption[Multi-modal data acquisition and forest structure derivation framework]{\textbf{Multi-modal data acquisition and forest structure derivation framework.} (a) RGBI orthophotos and LiDAR are acquired at different times. (b) Inputs are spatially aligned in 2 km \(\times\) 2 km tiles. (c) LiDAR provides privileged structural supervision (CHM, PAI, FHD, PAD) for teacher training, while the deployed student uses RGBI only.}
    \label{fig:fig1}
\end{figure}

Forest structure mapping is a 3D perception problem that underpins carbon accounting, biodiversity monitoring, and ecosystem analysis \cite{kamoske2019leaf,lim2003lidar}. Airborne LiDAR remains the reference source for fine-scale canopy geometry, but repeated campaigns are costly and typically infrequent \cite{fassnacht2025forest,puletti2020lidar}. In contrast, very-high-resolution (VHR) aerial RGBI imagery is collected routinely (e.g., 20 cm DOP coverage in Germany \cite{BKG_DOP_DE}) (\autoref{fig:fig1}). This creates a clear computer vision question: how much 3D forest structure can be recovered from single-view optical imagery alone \cite{brandt2025high}?

Recent work has shown that monocular canopy height mapping is feasible. Two representative and directly comparable estimators are HRCHM \cite{tolan2024very} and DAC \cite{rege2024depth,yang2024depth}, both designed for CHM prediction from RGB imagery. These are important advances, but they remain height-only methods and do not explicitly target richer vertical structure descriptors.

Specifically, HRCHM couples a DINOv2-style representation with a dense prediction decoder for high-resolution CHM mapping \cite{tolan2024very,oquab2023dinov2}, while DAC adapts Depth Anything v2 representations to canopy height estimation with substantially lower model complexity \cite{rege2024depth,yang2024depth}. Together, they establish strong monocular CHM baselines, but neither is designed to predict multi-metric vertical structure.

That limitation is operationally important. Forest monitoring needs canopy height (CHM), but also metrics such as Plant Area Index (PAI) and Foliage Height Diversity (FHD), which reflect canopy layering and ecosystem function \cite{kamoske2019leaf}. These quantities are only weakly observable in a single nadir image, so direct RGBI-only supervision is underconstrained for robust transfer.

We therefore use learning with privileged information \cite{vapnik2015learning} and knowledge distillation \cite{hinton2015distilling,gou2021knowledge}. During training, a multi-modal teacher uses RGBI plus LiDAR-derived planar and vertical cues; during inference, a student runs on RGBI only. The design is motivated by cross-modal transfer results in autonomous driving, where LiDAR-aware teachers improve monocular students on geometric tasks \cite{chong2022monodistill,chen2022bevdistill,lan2022instance,li2023mkd,gao2024monofg}.

Our framework predicts CHM, PAI, and FHD at 20 cm resolution from RGBI input. We train and evaluate on public data \cite{sachsenLuftbildProdukteOffene} with geographically separated splits to test zero-shot regional transfer. The intended use is practical: frequent structure-map refreshes between LiDAR campaigns for operational workflows such as Digital Twin Germany \cite{bundDigitalTwin}.

This paper makes three contributions:
\begin{itemize}
    \item A LiDAR-to-RGBI teacher--student distillation framework for dense forest structure prediction.
    \item A multi-target monocular setup (CHM, PAI, FHD), extending beyond height-only inference.
    \item A CV-oriented evaluation against strong monocular CHM baselines.
\end{itemize}

\section{Data and Methods}

\section{Study Area}

Our experiments are conducted in Saxony (Germany), where forests cover roughly 28\% of the state (about 520,000 ha) and include both conifer-dominated and broadleaf systems \cite{eustaforEuropeanState,hering2005conversion}. This diversity makes the region suitable for testing both dense structural prediction and cross-region transfer behavior.

\begin{figure}[hbt!]
    \centering
    \includegraphics[width=1\linewidth]{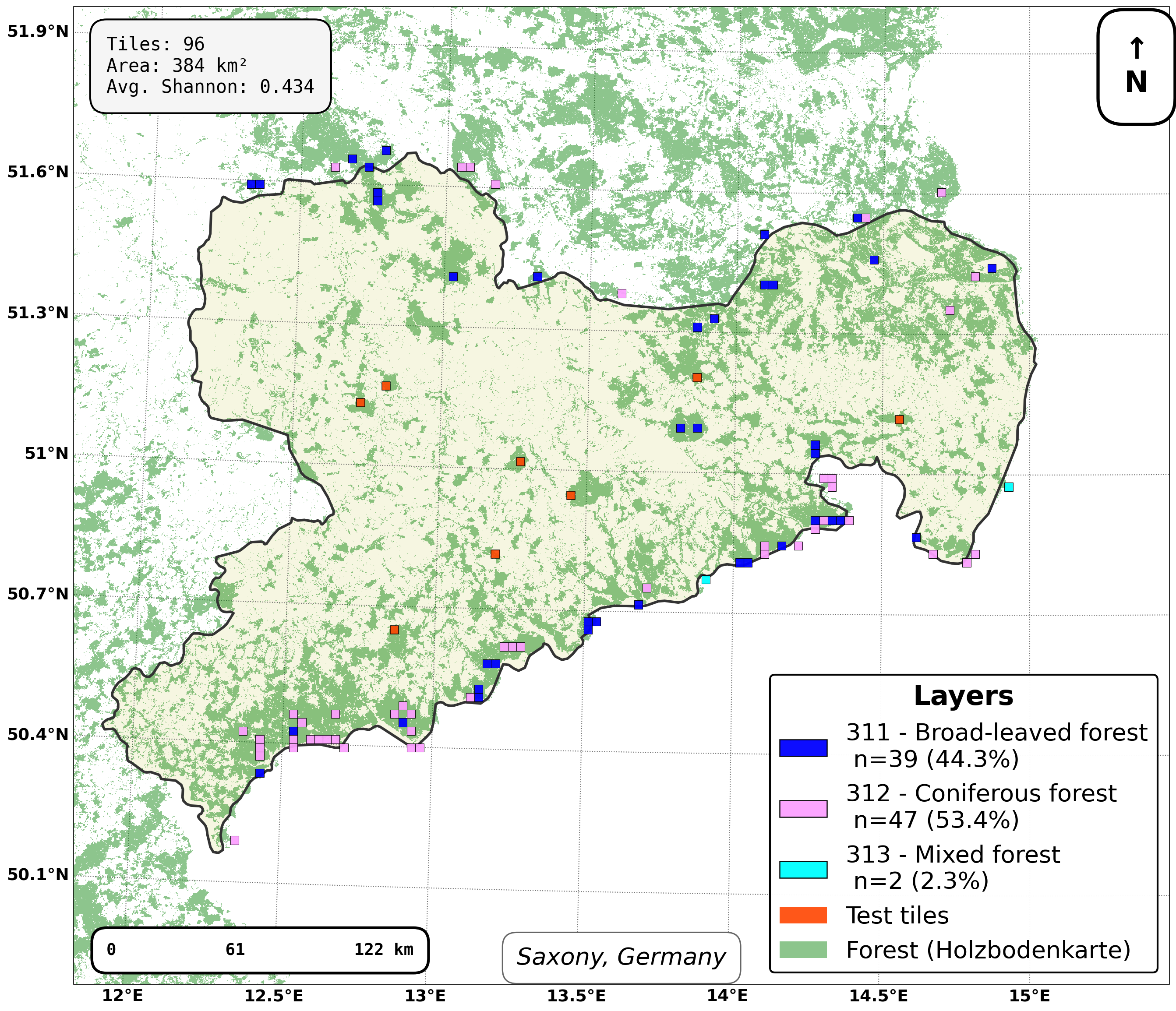}
    \caption[Study area and dataset spatial distribution across Saxony, Germany]{\textbf{Study area and dataset split.} Spatial distribution of the 96 RGBI--LiDAR tile pairs in Saxony and neighboring areas with geographically separated train/validation/test regions.}
    \label{fig:fig2}
\end{figure}

We use ninety-six paired RGBI--LiDAR tiles (2\,km $\times$ 2\,km each; 384\,km$^2$ total). To reduce spatial leakage, we enforce a geographic split into eighty training, eight validation, and eight test tiles, with test locations held out from training regions (\autoref{fig:fig2}). Sampling is stratified by CORINE forest composition \cite{EEA_CLC2018}, yielding a mix dominated by conifer and broadleaf classes with a smaller mixed-forest fraction.

\section{Data Sources}

\paragraph{Aerial Orthophotos (RGBI)} We use 20\,cm GSD four-band orthophotos (R, G, B, NIR) from GeoSN DOP products \cite{sachsenLuftbildProdukteOffene}. These images represent the operational modality that is available at high refresh rate and large area coverage under national orthophoto programs \cite{BKG_DOP_DE}. The selected tiles are primarily leaf-on (2022--2023).

\paragraph{Airborne LiDAR Point Clouds} Co-registered LiDAR (LAS/LAZ) is taken from GeoSN products \cite{sachsenFachlicheDetails}. Acquisition dates are not fully aligned with RGBI (mostly leaf-off LiDAR, typically 2017--2023), which reflects realistic deployment conditions rather than ideal co-acquisition. The LiDAR tiles have a mean point density of 15.82 points/m², ranging from 0.98 to 24.47 points/m².

\section{Data Preprocessing and Spatial Alignment}

\begin{figure}[!htb]
    \centering
    \includegraphics[width=1\linewidth]{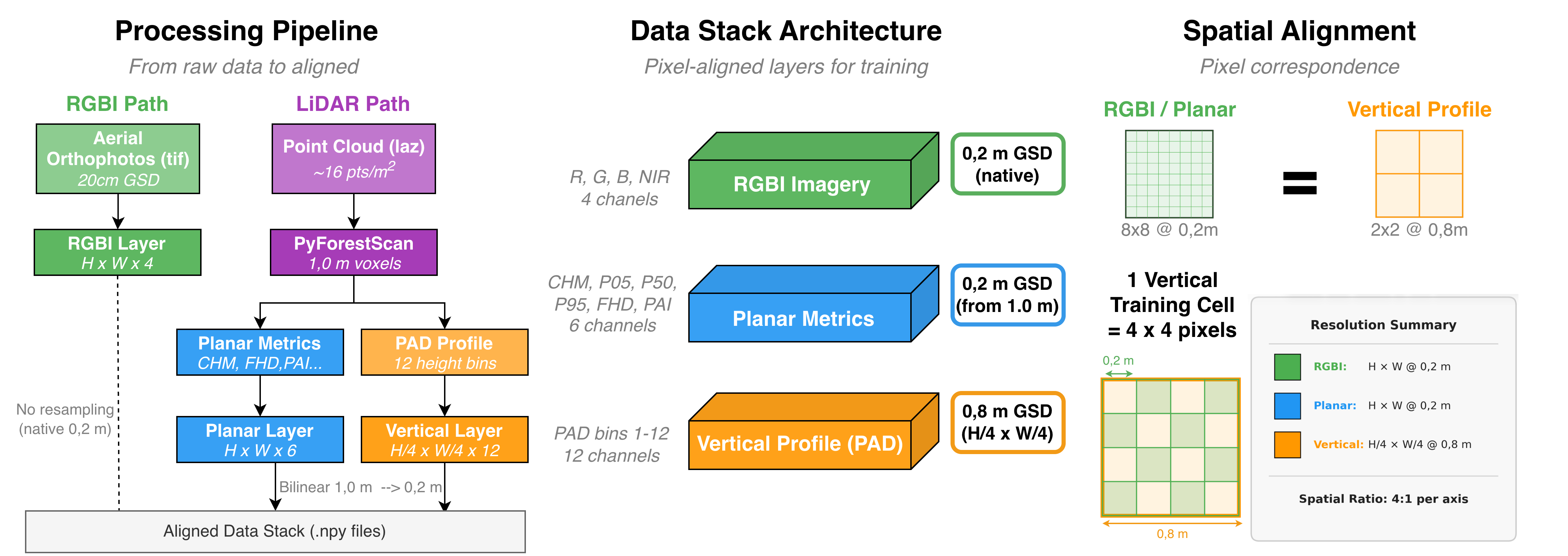}
    \caption[Data preprocessing pipeline and multi-resolution spatial alignment strategy.]{\textbf{Preprocessing and alignment overview.} RGBI and LiDAR products are transformed into pixel-aligned multi-modal training tensors for distillation.}
    \label{fig:fig3}
\end{figure}

Training uses 224$\times$224 patches (about 45\,m $\times$ 45\,m at 20\,cm GSD) sampled from forested areas. This patch-based setup keeps training tractable while preserving local canopy structure, as shown in \autoref{fig:fig3}. For each patch we prepare:
\begin{itemize}
    \item RGBI imagery (4 channels),
    \item planar LiDAR-derived targets (CHM, PAI, FHD) using equations from \autoref{tab:canopy_metrics},
    \item vertical Plant Area Density (PAD) profiles.
\end{itemize}

LiDAR point clouds are processed with PyForestScan \cite{percival2025pyforestscan}. Metrics are first computed on a 1\,m grid, then aligned to the RGBI reference grid (EPSG:25833), and resampled to 20\,cm where required \cite{candan2023udi,ressl2012applying}. PAD profiles are retained as vertical supervision and stored at lower effective spatial resolution for memory-efficient training \cite{harris2020array}. A validity mask removes invalid pixels before loss computation so supervision is restricted to spatially reliable regions.

A sample pre-processing log file is provided in the supplementary material.

\begin{table}[!htb]
    \centering
    \caption{Common canopy structure metrics and their equations\cite{kamoske2019leaf,percival2025pyforestscan}.}
    \label{tab:baselines}
    \label{tab:canopy_metrics}
    \begin{tabular}{|l|l|}
        \toprule
        \textbf{Metric} & \textbf{Equation} \\
        \midrule
        CHM - planar & 
            $\text{H}_{\text{canopy}} = \max(\text{HAG}_{\text{points}})$ \\
        \midrule
        PAD - vertical & 
            $ \text{PAD}_{i-1,i} = \ln \left( \frac{S_e}{S_t} \right) \frac{1}{k \Delta z} $ \\
        \midrule
        PAI - planar & 
            $ \text{PAI} = \sum_{i=1}^{n} \text{PAD}_{i-1,i} $ \\
        \midrule
        FHD - planar & 
            $ \text{FHD} = - \sum_{i=1}^{n} p_i \ln(p_i) $ \\
        \midrule
        P05, P50, P95 & 
            $P_{xx}$ is the height below which $xx\%$ of all points lie. \\
        \bottomrule
    \end{tabular}
\end{table}

\section{Model Architecture}

\begin{figure}
    \centering
    \includegraphics[width=1\linewidth]{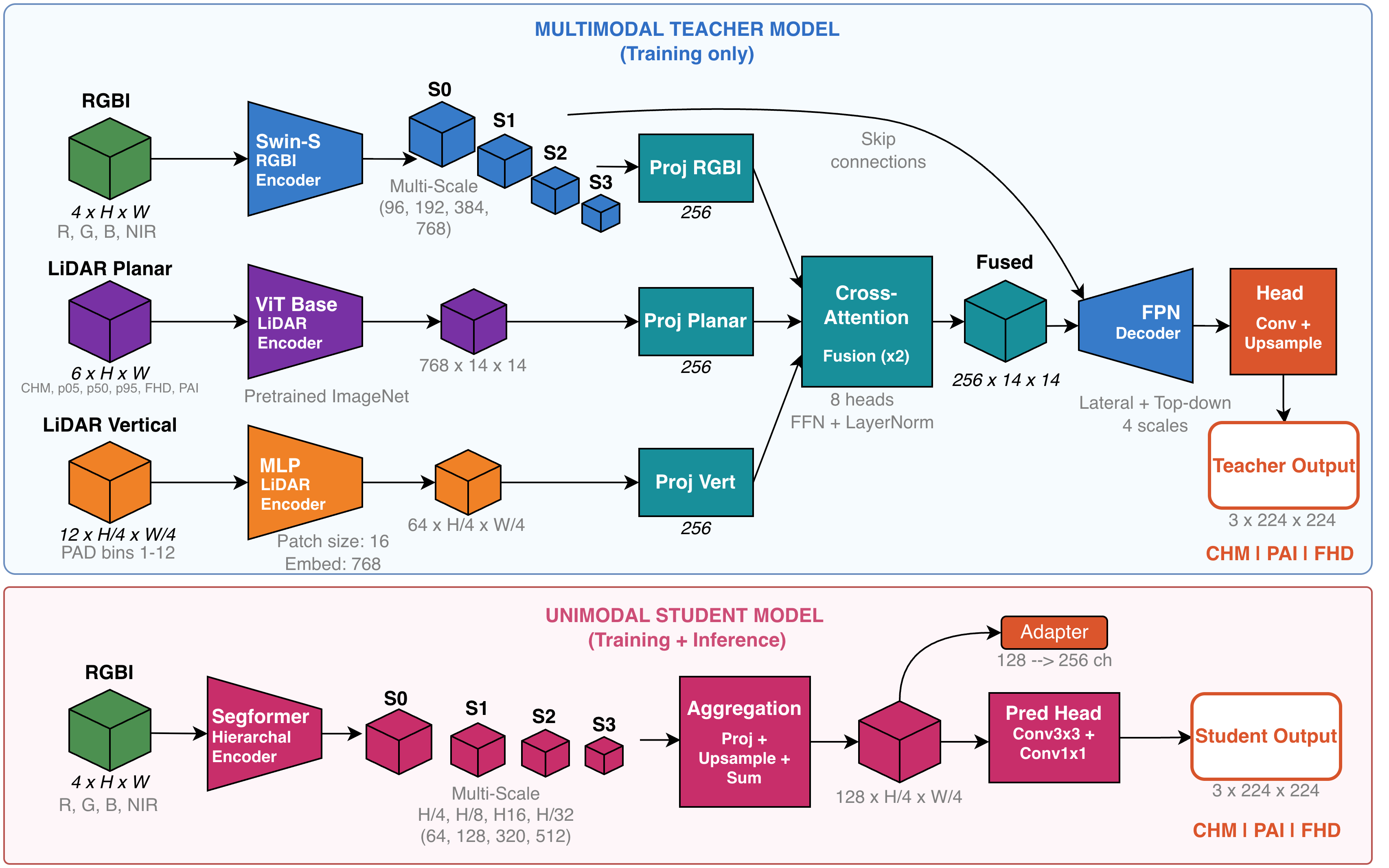}
    \caption[Multi-modal teacher and unimodal student network architectures for forest structure prediction.]{\textbf{Teacher--student architecture.} A LiDAR-aware teacher supervises an RGBI-only student for CHM/PAI/FHD prediction.}
    \label{fig:fig4}
\end{figure}

As shown in \autoref{fig:fig4}, we train a multi-modal teacher and an RGBI-only student. Both output dense CHM, PAI, and FHD maps.

\subsection{Teacher Model (Multi-Branch Fusion)}

\begin{figure}[!htb]
    \centering
    \includegraphics[width=1\linewidth]{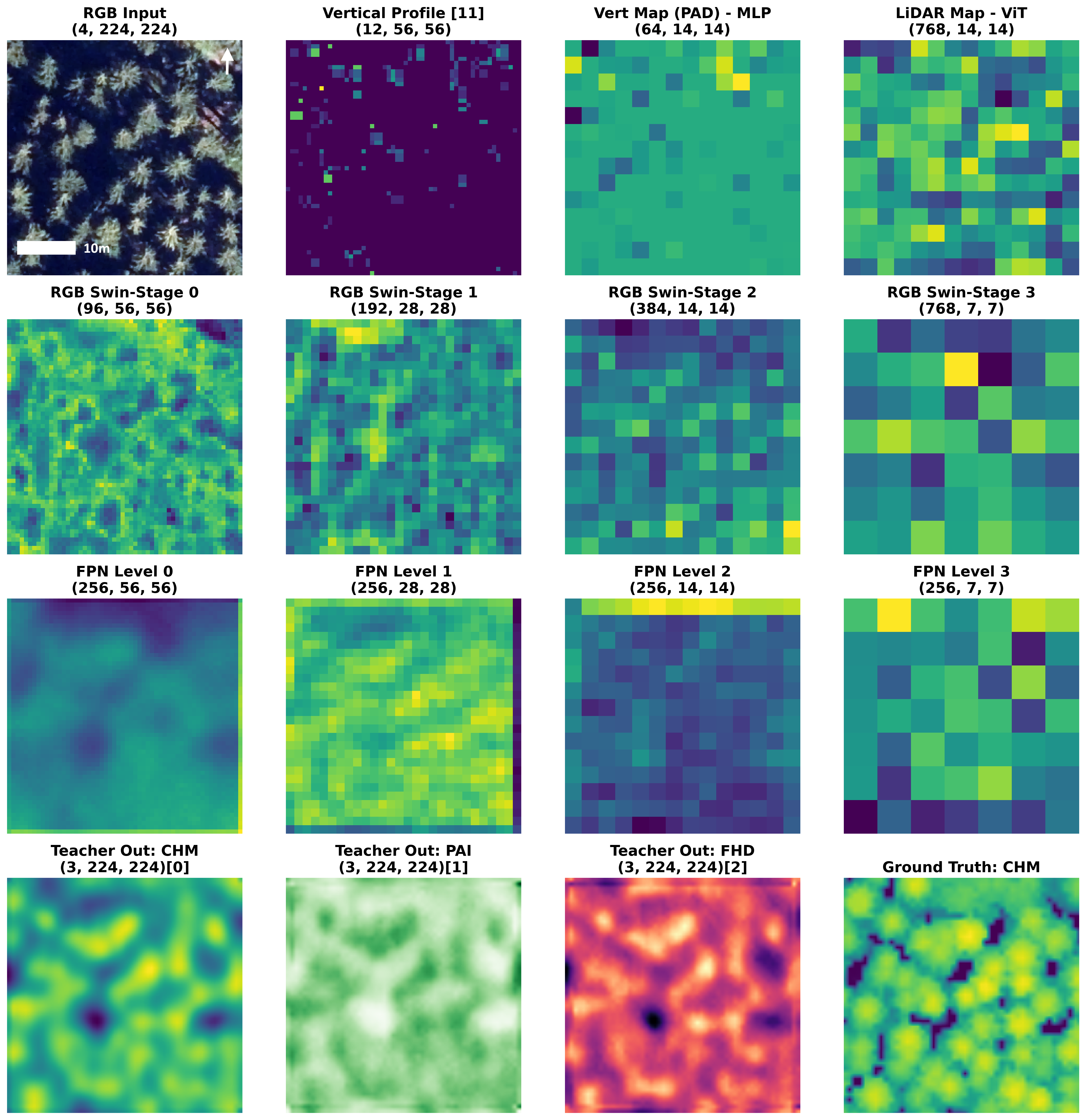}
    \caption[Teacher model feature hierarchy and multi-modal fusion.]{\textbf{Teacher feature flow.} Multi-modal feature extraction, fusion, and decoding in the teacher network.}
    \label{fig:fig5}
\end{figure}

The teacher fuses three streams: RGBI features (Swin), planar LiDAR features (ViT), and vertical PAD features (MLP encoder), followed by cross-attention fusion and an FPN decoder \cite{liu2021swin,dosovitskiy2020image,li2024crossfuse,lin2017feature}. This model is used as a privileged-information supervisor during distillation. \autoref{fig:fig5} illustrates the teacher feature maps and fusion flow.

\subsection{Student Model (Monocular RGBI Estimator)}

\begin{figure}[!htb]
    \centering
    \includegraphics[width=1\linewidth]{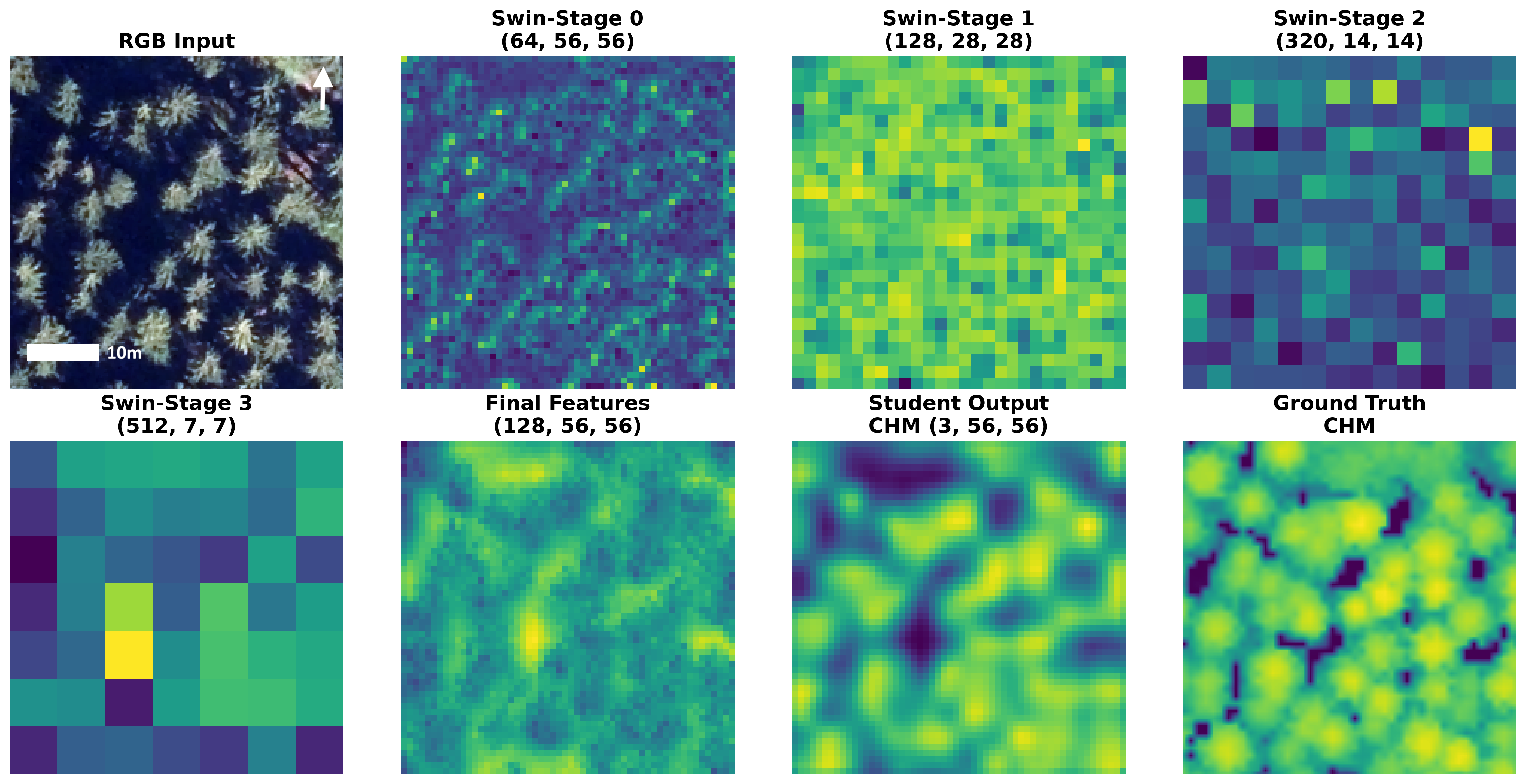}
    \caption[Student model feature hierarchy and CHM prediction]{\textbf{Student feature flow.} Multi-scale RGBI encoding and dense regression in the RGBI-only student.}
    \label{fig:fig6}
\end{figure}

The student uses SegFormer (MiT-B2/B5) with four-channel RGBI input and a lightweight regression head \cite{xie2021segformer}. Multi-scale encoder features are fused, then mapped to the three output metrics. A 1$\times$1 adapter projects student features into the teacher fusion space for feature distillation. \autoref{fig:fig6} illustrates the feature hierarchy and flow.

\[
\text{Student}(\text{RGBI}) \rightarrow [\tilde{\text{CHM}},\tilde{\text{PAI}},\tilde{\text{FHD}}].
\]

\section{Knowledge Distillation \& Training}

\paragraph{Training protocol} Training is two-stage: (1) train the teacher on multi-modal inputs, then (2) freeze the teacher and train the student with supervised and distillation losses.

\subsection{Stage 1: Teacher}

Teacher optimization uses robust regression over CHM/PAI/FHD plus a CHM gradient term:
\begin{equation}
\mathcal{L}_{\mathrm{teacher}}
=
\sum_{c\in\{\mathrm{CHM,PAI,FHD}\}}
\mathcal{L}_{\mathrm{robust}}(Y_c^{t}, Y_c)
+
\lambda_{\mathrm{grad}}\,\mathcal{L}_{\mathrm{grad}}(Y_{\mathrm{CHM}}^{t}, Y_{\mathrm{CHM}}).
\end{equation}
Here, $\mathcal{L}_{\mathrm{robust}}$ is a masked Smooth L1 (Huber) regression loss applied per channel with equal weighting across CHM/PAI/FHD. It is quadratic for small residuals and linear for large residuals, improving robustness to outliers while preserving fine-structure sensitivity.

\begin{equation}
\mathcal{L}_{\mathrm{grad}}
=
\left\lVert \nabla_x Y_{\mathrm{CHM}}^{t} - \nabla_x Y_{\mathrm{CHM}} \right\rVert_1
+
\left\lVert \nabla_y Y_{\mathrm{CHM}}^{t} - \nabla_y Y_{\mathrm{CHM}} \right\rVert_1 ,
\end{equation}
where $\nabla_x$ and $\nabla_y$ are finite-difference gradients. We include this term because CHM contains the strongest spatial gradient information among the targets; matching CHM gradients improves edge and transition fidelity (e.g., canopy boundaries) and reduces over-smoothing \cite{tong2023functional}. We use $\lambda_{\mathrm{grad}}=0.1$.
The teacher is optimized with AdamW (lr=$1\times10^{-4}$, weight decay=$1\times10^{-4}$).

\subsection{Stage 2: Student with KD}

Student training combines supervised output fitting, output distillation from the frozen teacher, feature matching, and vertical-proxy alignment:
\begin{equation}
\begin{aligned}
L_{\mathrm{out}}  &= \left\lVert Y^{s} - Y \right\rVert_1, \\
L_{\mathrm{KD}}   &= \mathrm{SmoothL1}(Y^{s}, Y^{t}), \\
L_{\mathrm{feat}} &= \left\lVert \mathrm{Proj}(F_{\mathrm{fused}}^{s}) - F_{\mathrm{fused}}^{t} \right\rVert_2^2, \\
L_{\mathrm{vert}} &= \left\lVert \mathrm{Proj}(F_{\mathrm{fused}}^{s}\!\downarrow) - F_{\mathrm{fused}}^{t} \right\rVert_2^2 .
\end{aligned}
\end{equation}
\begin{equation}
\mathcal{L}_{\mathrm{student}} =
 w_{\mathrm{sup}}L_{\mathrm{out}}
+ w_{\mathrm{KD}}L_{\mathrm{KD}}
+ w_{\mathrm{feat}}L_{\mathrm{feat}}
+ w_{\mathrm{vert}}L_{\mathrm{vert}}.
\end{equation}
We keep $w_{\mathrm{sup}}=1$ and use auxiliary weights in a low range (0--0.5). The selected setting is $w_{\mathrm{KD}}=0.5$, $w_{\mathrm{feat}}=0.1$, and $w_{\mathrm{vert}}=0.1$, with KD warm-up ($w_{\mathrm{KD}}=0$ in early epochs).
The student is optimized with AdamW (lr=$2\times10^{-4}$, weight decay=$1\times10^{-4}$).

\paragraph{Teacher symbols} $Y$: LiDAR-derived targets; $Y^{t}$: teacher predictions; $\mathcal{L}_{\mathrm{robust}}$: masked Smooth L1 regression; $\mathcal{L}_{\mathrm{grad}}$: CHM gradient consistency; $\lambda_{\mathrm{grad}}$: gradient-loss weight.
\paragraph{Student symbols} $Y^{s}$: student predictions; $F_{\mathrm{fused}}^{t},F_{\mathrm{fused}}^{s}$: teacher/student fused features; $\downarrow$: downsampling to teacher fusion scale; $\mathrm{Proj}$: $1{\times}1$ adapter; $w_{\cdot}$: student loss weights.

We use standard geometric/color augmentation (random patch sampling/cropping only) and spatially separated train/validation/test splits (Tile-pairs:80/8/8). A sample training log file is provided in the supplementary material.

\section{Evaluation}

\subsection{Metrics}

On validation tiles, we report MAE, RMSE, Bias, and $R^2$ to track fit quality and systematic error during model selection. On held-out test tiles, we additionally report MedAE, Pearson $R$, IoU, F1, and rMAE to characterize zero-shot transfer in both continuous-value accuracy and spatial agreement.

\subsection{Ablations}

We run comprehensive targeted ablations (\autoref{tab:ablation_studies}) to measure the contribution of distillation, fusion, backbone scale, and optimization choices \cite{sheikholeslami2019ablation}. The model weights for each ablation are provided in the supplementary material for reproducibility and further analysis.

\begin{table}[!htb]
    \centering
    \caption{Ablation studies for model architecture and training parameters}
    \label{tab:ablation_studies}
    \begin{tabularx}{\textwidth}{lXX}
        \toprule
        \textbf{Study} & \textbf{What to remove/test} & \textbf{Why} \\
        \midrule
        No distillation loss & Remove KD loss $\to$ only RGBI supervision. & Does distillation actually help? \\
        \midrule
        Asymmetric learning & 1. Train teacher for fewer epochs than student. \newline 2. Train teacher for more epochs than student. & Does symmetry matter in learning representations? \\
        \midrule
        \parbox[t]{3cm}{Without multimodal fusion\\(no cross-attention)} & Remove multi-modal fusion. & Does multi-stage feature fusion via cross-attention matter? \\
        \midrule
        Batch size sensitivity & Try batch size 8 vs 64. & Important for generalization vs. memory. \\
        \midrule
        Backbone size & Try MiT-B2 vs B5. & Smaller vs larger (impact on capacity). \\
        \bottomrule
    \end{tabularx}
\end{table}

\subsection{SOTA CHM Baselines}

For CHM, we compare against two monocular SOTA families: HRCHM \cite{tolan2024very} and DAC \cite{rege2024depth,yang2024depth}. We use the model weights provided with the original papers for inference on test tiles.

Following the setup in the full paper, HRCHM is evaluated in both native 20\,cm and scale-matched 60\,cm variants, and DAC outputs are converted to metric CHM before scoring. All baseline outputs are then reprojected to the same 20\,cm evaluation grid for pixel-wise comparison.

\subsection{Inference Pipeline}

At deployment, the student performs sliding-window inference over full RGBI tiles and stitches predictions into seamless georeferenced CHM/PAI/FHD rasters.

\section{Results}

\section{Quantitative Performance}

All results are provided as tables in the supplementary material.

\subsection{Core Outcomes}

We report compact headline metrics here, then interpret what they imply for transfer behavior.

\paragraph{Validation summary}
On validation data, the student retained most of the teacher signal for CHM (teacher/student MAE: 3.88/4.95 m; $R^2$: 0.69/0.61) while reducing bias magnitude. For FHD, teacher and student were nearly identical ($R^2=0.54$ for both), indicating that this target transfers stably under the training distribution. For PAI, the student outperformed the teacher (MAE: 0.48 vs 0.59; $R^2$: 0.31 vs 0.05), which suggests that distillation helped suppress teacher-specific noise and regularize the final predictor.

\paragraph{Zero-shot test summary}
On the geographically held-out test set, CHM generalized best. It reached MAE 5.81 m, MedAE 4.17 m, $R=0.713$, $R^2=0.509$, IoU 0.870, and F1 0.930. The main residual error mode is underestimation in taller canopies (bias -2.57 m), but overall spatial structure remains robust.

FHD and PAI transferred less reliably (FHD: MAE 0.42, $R=0.391$, rMAE 29.7\%; PAI: MAE 0.46, $R=0.361$, rMAE 44.0\%). This gap relative to CHM is consistent with domain shift for vertically complex descriptors: height cues transfer more readily across regions than canopy-layer composition cues.

\begin{figure}[!htb]
    \centering
    \includegraphics[width=1\linewidth]{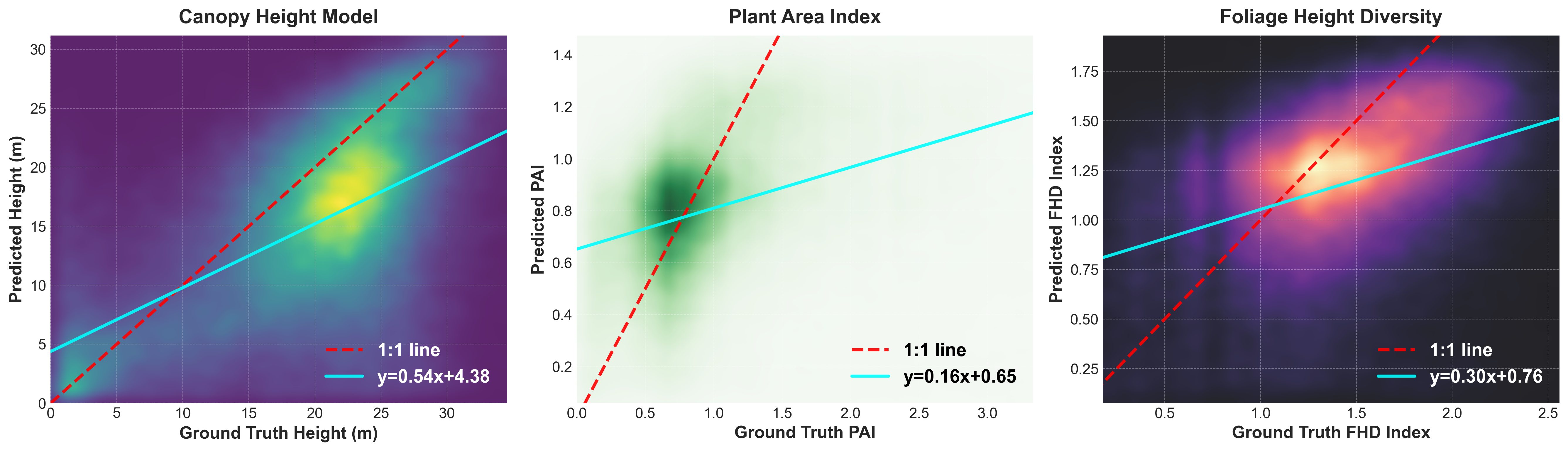}
    \caption[Ground truth versus predicted forest structure metrics on test tiles]{\textbf{Quantitative agreement on held-out test tiles.} CHM shows the strongest correlation; PAI and FHD remain harder under zero-shot transfer.}
    \label{fig:fig8}
\end{figure}

\subsection{Ablation Summary}
Ablations reveal three stable effects. First, multi-modal fusion is a primary driver: removing it degrades all targets, with the largest relative drop for FHD. Second, asymmetric distillation with synchronized teacher/student schedules performs best; extended student training against a weak teacher leads to instability. Third, MiT-B2 is more reliable than MiT-B5 in this data regime, indicating that capacity must match supervision density rather than simply scale upward. The full ablation results table is provided in the supplementary material.

\subsection{SOTA CHM Comparison}
Against HRCHM and DAC baselines, our model achieved the best CHM accuracy on the same test tiles (\autoref{tab:comparison_sota}): MAE 5.81 m versus 8.14 m (HRCHM-60cm), 9.88 m (HRCHM-full-res), and 10.84 m (DAC-B). Correlation was also strongest ($R=0.713$ vs 0.652/0.452/0.166), with lower systematic underestimation.

These results support the central claim: LiDAR-informed distillation transfers stronger structural priors than purely monocular baselines in this setting, even when all methods are evaluated on the same 20\,cm target grid.

Importantly, this is a stringent comparison: the baselines are strong CHM-focused monocular methods, whereas our model is trained for joint CHM/PAI/FHD prediction and evaluated in a geographically held-out regime.

\begin{table}[h]
\centering
\caption{Comparison of evaluation metrics with state-of-the-art methods on CHM prediction.}
\label{tab:comparison_sota}
\resizebox{\textwidth}{!}{%
\begin{tabular}{|l|c|c|c|c|c|c|c|}
\hline
\textbf{Model} & \textbf{MAE (m)} & \textbf{MedAE (m)} & \textbf{RMSE (m)} & \textbf{Bias (m)} & \textbf{R} & \textbf{IoU} & \textbf{F1} \\
\hline
FSKD (MiT-B2/b64) & $5,81 \pm 1,44$ & $4,17 \pm 1,62$ & 7,90 & -2,57 & 0,713 & 0,870 & 0,930 \\
\hline
HRCHM-aerial (Full Res) & $9,88 \pm 2,49$ & $8,83 \pm 3,93$ & 12,89 & -8,55 & 0,452 & 0,733 & 0,842 \\
\hline
HRCHM-aerial (60cm) & $8,14 \pm 2,64$ & $7,46 \pm 3,01$ & 10,34 & -6,94 & 0,652 & 0,834 & 0,907 \\
\hline
DAC-B & $10,84 \pm 2,38$ & $10,99 \pm 3,09$ & 12,56 & -6,49 & 0,166 & 0,801 & 0,888 \\
\hline
\end{tabular}
}
\end{table}

\section{Qualitative Performance}

\begin{figure}[!htb]
    \centering
    \includegraphics[width=1\linewidth]{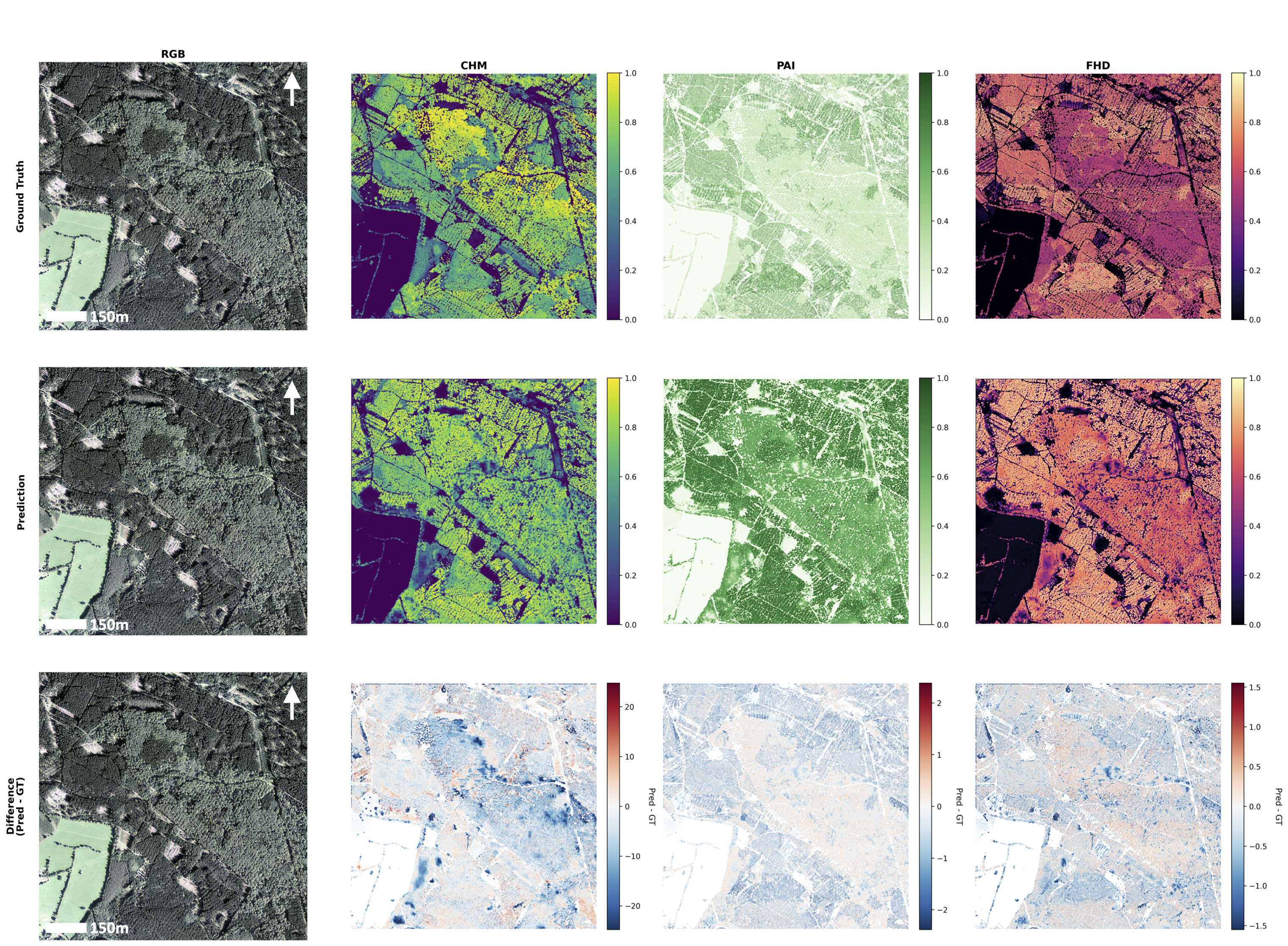}
    \caption[Qualitative evaluation of student predictions on a representative test tile]{\textbf{Representative tile.} Normalised (for comparison) student CHM/PAI/FHD maps are spatially coherent with LiDAR-derived references, with strongest agreement for CHM.}
    \label{fig:fig10}
\end{figure}

\begin{figure}[!htb]
    \centering
    \includegraphics[width=1\linewidth]{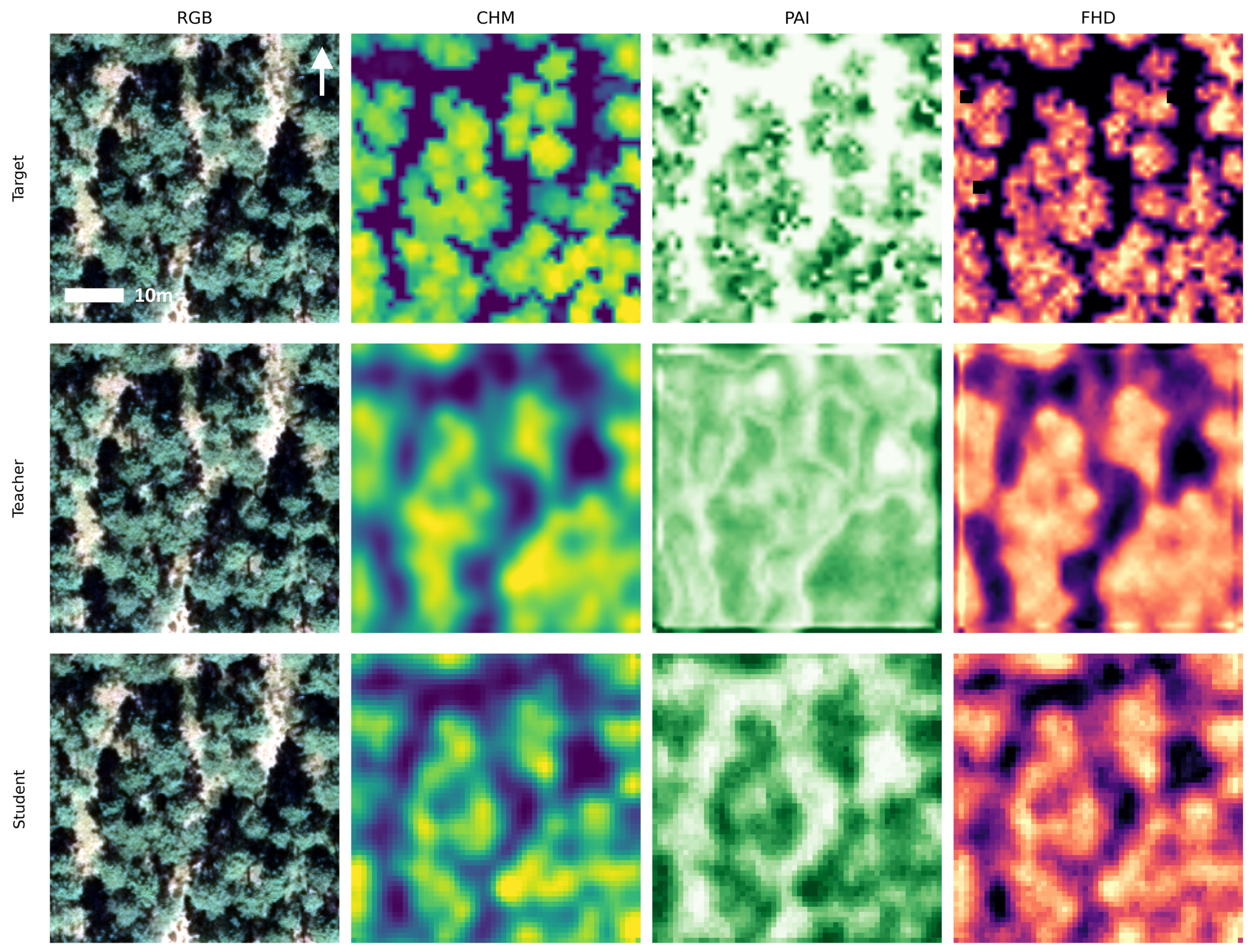}
    \caption[Comparison of ground truth, teacher, and student predictions on a broadleaved forest patch]{\textbf{Broadleaved patch comparison.} Ground truth (top), teacher (middle), and student (bottom) show close CHM agreement with clear crown delineation; PAI/FHD preserve major spatial gradients but appear smoother.}
    \label{fig:fig11}
\end{figure}

\begin{figure}[!htb]
    \centering
    \includegraphics[width=1\linewidth]{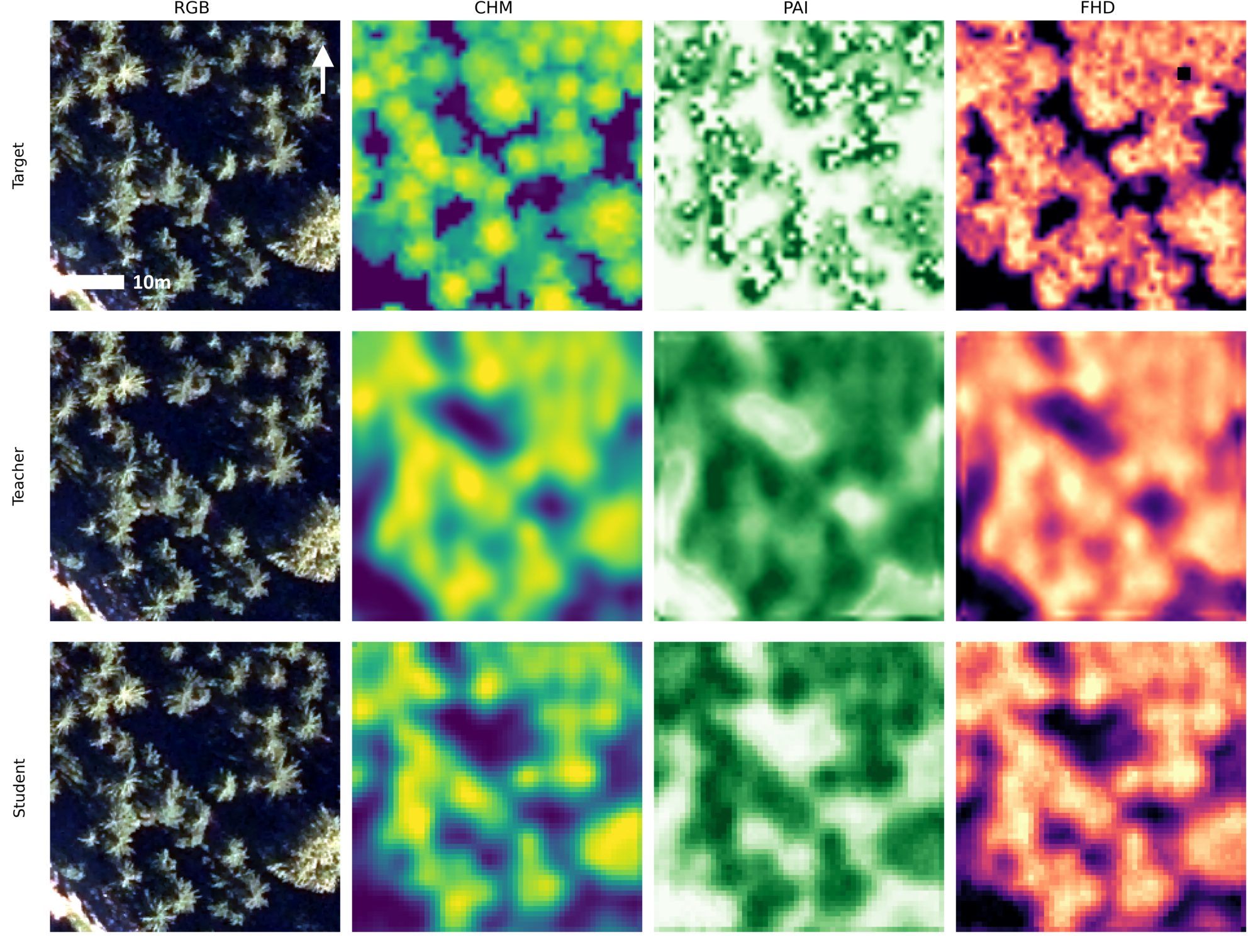}
    \caption[Comparison of ground truth, teacher, and student predictions on a coniferous forest patch (45×45 m)]{\textbf{Coniferous patch comparison.} Ground truth (top), teacher (middle), and student (bottom) show strong CHM transfer; PAI/FHD remain directionally correct with compressed amplitude.}
    \label{fig:fig12}
\end{figure}

Qualitative maps match the quantitative pattern. CHM predictions preserve stand boundaries and canopy gradients robustly, while PAI/FHD preserve broad structure but compress dynamic range in difficult regions. \autoref{fig:fig10} illustrates tile-scale behavior, whereas \autoref{fig:fig11} and \autoref{fig:fig12} highlights fine-structure transfer at patch scale for different forest types.

\section{Discussion}

\paragraph{What works}
The teacher--student setup reliably transfers LiDAR-informed geometry into an RGBI-only model. The strongest and most stable signal is CHM, where both quantitative and qualitative evidence indicates usable cross-region performance at 20 cm output resolution.

\paragraph{Where it breaks}
PAI and FHD are more sensitive to regional canopy composition and acquisition mismatch. Their weaker transfer indicates that monocular cues alone are insufficient for universally calibrated vertical-structure inference without broader training coverage, stronger priors, or explicit adaptation \cite{flynn2024uav,tan2024exploring,burns2024multi}.

\paragraph{Operational implications}
The student is lightweight at inference and can refresh structural layers whenever new RGBI imagery is available, which is useful for workflows such as Digital Twin Germany \cite{bundDigitalTwin}. In practice, this is best treated as a LiDAR-complement strategy: frequent optical updates between less frequent LiDAR campaigns, with periodic recalibration to maintain vertical-metric fidelity.

\begin{figure}[!htb]
    \centering
    \includegraphics[width=1\linewidth]{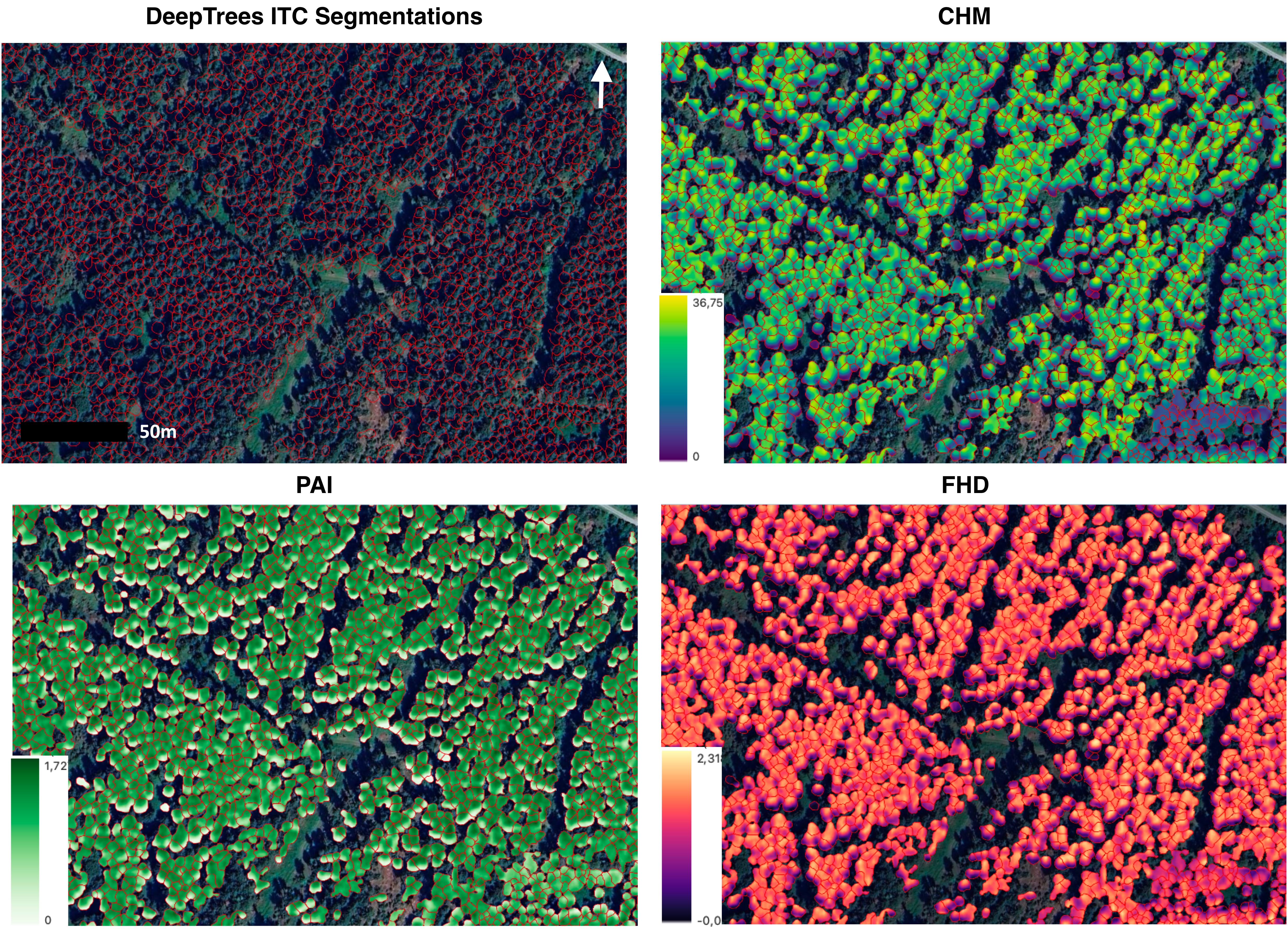}
    \caption[Individual tree crown segmentation applied to all predicted forest structure metrics]{\textbf{Metric-wise ITC masking.} Tree-crown masks derived with DeepTrees \cite{Khan2025} can be applied consistently to CHM, PAI, and FHD outputs for per-tree summaries.}
    \label{fig:fig14}
\end{figure}

\paragraph{Near-term extensions}
Priority next steps are: (1) expand training coverage across more Saxony tiles/years/seasons (leaf-on vs. leaf-off), (2) add uncertainty estimation for map-level confidence, (3) test additional modalities (e.g., Synthetic Aperture Radar - SAR or multi-temporal RGBI) and feature extractors (e.g. DINO), and (4) integrate crown-level products for downstream workflows \cite{alidoost20192d,chen2023integrating,lin2024individual,Khan2025,xiang2025forestformer3d}, as demonstrated in \autoref{fig:fig14}.

\section{Conclusion}

This work shows that LiDAR-to-RGBI knowledge distillation can produce a practical monocular forest-structure estimator with strong CHM performance and useful first-order PAI/FHD signals. The key result is operational: LiDAR-informed structure can be propagated to routine aerial imagery without LiDAR at inference time.

The method contributes: (1) a cross-modal teacher--student pipeline for CHM/PAI/FHD at 20 cm, (2) stable gains from privileged multi-modal fusion over RGBI-only training, and (3) a deployment path for frequent large-area updates between LiDAR acquisitions.

The remaining gap is generalization of vertical structure indices (PAI/FHD). Closing that gap likely requires broader regional training, uncertainty-aware prediction, and targeted domain adaptation rather than architectural scaling alone. Beyond direct mapping, FSKD outputs can also serve as inputs for downstream deep learning tasks, including tree-crown segmentation, tree vitality estimation, species classification, and biomass estimation.

Overall, FSKD establishes a practical computer-vision blueprint for converting routine RGBI imagery into actionable 3D forest structure at scale.
\clearpage
\bibliographystyle{splncs04}
\bibliography{main}

@article{tolan2024very,
  title={Very high resolution canopy height maps from RGB imagery using self-supervised vision transformer and convolutional decoder trained on aerial lidar},
  author={Tolan, Jamie and Yang, Hung-I and Nosarzewski, Benjamin and Couairon, Guillaume and Vo, Huy V and Brandt, John and Spore, Justine and Majumdar, Sayantan and Haziza, Daniel and Vamaraju, Janaki and others},
  journal={Remote Sensing of Environment},
  volume={300},
  pages={113888},
  year={2024},
  publisher={Elsevier}
}

@article{lim2003lidar,
  title={LiDAR remote sensing of forest structure},
  author={Lim, Kevin and Treitz, Paul and Wulder, Michael and St-Onge, Beno{\^\i}t and Flood, Martin},
  journal={Progress in physical geography},
  volume={27},
  number={1},
  pages={88--106},
  year={2003},
  publisher={Sage Publications Sage CA: Thousand Oaks, CA}
}

@article{hinton2015distilling,
  title={Distilling the knowledge in a neural network},
  author={Hinton, Geoffrey and Vinyals, Oriol and Dean, Jeff},
  journal={arXiv preprint arXiv:1503.02531},
  year={2015}
}

@article{gou2021knowledge,
  title={Knowledge distillation: A survey},
  author={Gou, Jianping and Yu, Baosheng and Maybank, Stephen J and Tao, Dacheng},
  journal={International journal of computer vision},
  volume={129},
  number={6},
  pages={1789--1819},
  year={2021},
  publisher={Springer}
}

@article{dosovitskiy2020image,
  title={An image is worth 16x16 words: Transformers for image recognition at scale},
  author={Dosovitskiy, Alexey},
  journal={arXiv preprint arXiv:2010.11929},
  year={2020}
}

@article{lin2024individual,
  title={Individual Tree Crown Delineation Using Airborne LiDAR Data and Aerial Imagery in the Taiga--Tundra Ecotone},
  author={Lin, Yuanyuan and Li, Hui and Jing, Linhai and Ding, Haifeng and Tian, Shufang},
  journal={Remote Sensing},
  volume={16},
  number={21},
  pages={3920},
  year={2024},
  publisher={MDPI}
}

@article{vapnik2015learning,
  title={Learning using privileged information: Similarity control and knowledge transfer.},
  author={Vapnik, Vladimir and Izmailov, Rauf and others},
  journal={J. Mach. Learn. Res.},
  volume={16},
  number={1},
  pages={2023--2049},
  year={2015}
}

@article{li2023mkd,
  title={MKD-cooper: Cooperative 3D object detection for autonomous driving via multi-teacher knowledge distillation},
  author={Li, Zhiyuan and Liang, Huawei and Wang, Hanqi and Zhao, Mingzhuo and Wang, Jian and Zheng, Xiaokun},
  journal={IEEE Transactions on Intelligent Vehicles},
  volume={9},
  number={1},
  pages={1490--1500},
  year={2023},
  publisher={IEEE}
}

@article{lan2022instance,
  title={Instance, scale, and teacher adaptive knowledge distillation for visual detection in autonomous driving},
  author={Lan, Qizhen and Tian, Qing},
  journal={IEEE Transactions on Intelligent Vehicles},
  volume={8},
  number={3},
  pages={2358--2370},
  year={2022},
  publisher={IEEE}
}

@article{gao2024monofg,
  title={MonoFG: Monocular 3D object detection with knowledge distillation for human-centric autonomous driving systems},
  author={Gao, Honghao and Yu, Xinxin and Xu, Yueshen and Ran, Qionghuizi and Hussain, Walayat},
  journal={ACM Transactions on Autonomous and Adaptive Systems},
  year={2024},
  publisher={ACM New York, NY}
}

@article{percival2025pyforestscan,
  title={PyForestScan: A Python library for calculating forest structural metrics from lidar point cloud data},
  author={Percival, Joseph Emile Honour and Leamon, Benjamin Palsa},
  journal={Journal of Open Source Software},
  volume={10},
  number={106},
  pages={7314},
  year={2025}
}

@article{kamoske2019leaf,
  title={Leaf area density from airborne LiDAR: Comparing sensors and resolutions in a temperate broadleaf forest ecosystem},
  author={Kamoske, Aaron G and Dahlin, Kyla M and Stark, Scott C and Serbin, Shawn P},
  journal={Forest Ecology and Management},
  volume={433},
  pages={364--375},
  year={2019},
  publisher={Elsevier}
}

@inproceedings{rege2024depth,
  title={Depth Any Canopy: Leveraging Depth Foundation Models for Canopy Height Estimation},
  author={Rege Cambrin, Daniele and Corley, Isaac and Garza, Paolo},
  booktitle={European Conference on Computer Vision},
  pages={71--86},
  year={2024},
  organization={Springer}
}

@article{fassnacht2025forest,
  title={Forest practitioners’ requirements for remote sensing-based canopy height, wood-volume, tree species, and disturbance products},
  author={Fassnacht, Fabian Ewald and Mager, Christoph and Waser, Lars T and Kanjir, Ur{\v{s}}a and Sch{\"a}fer, Jannika and Buhvald, Ana Poto{\v{c}}nik and Shafeian, Elham and Schiefer, Felix and Stan{\v{c}}i{\v{c}}, Liza and Immitzer, Markus and others},
  journal={Forestry: An International Journal of Forest Research},
  volume={98},
  number={2},
  pages={233--252},
  year={2025},
  publisher={Oxford University Press}
}

@article{brandt2025high,
  title={High-resolution sensors and deep learning models for tree resource monitoring},
  author={Brandt, Martin and Chave, Jerome and Li, Sizhuo and Fensholt, Rasmus and Ciais, Philippe and Wigneron, Jean-Pierre and Gieseke, Fabian and Saatchi, Sassan and Tucker, CJ and Igel, Christian},
  journal={Nature Reviews Electrical Engineering},
  volume={2},
  number={1},
  pages={13--26},
  year={2025},
  publisher={Nature Publishing Group UK London}
}

@article{oquab2023dinov2,
  title={Dinov2: Learning robust visual features without supervision},
  author={Oquab, Maxime and Darcet, Timoth{\'e}e and Moutakanni, Th{\'e}o and Vo, Huy and Szafraniec, Marc and Khalidov, Vasil and Fernandez, Pierre and Haziza, Daniel and Massa, Francisco and El-Nouby, Alaaeldin and others},
  journal={arXiv preprint arXiv:2304.07193},
  year={2023}
}

@article{yang2024depth,
  title={Depth anything v2},
  author={Yang, Lihe and Kang, Bingyi and Huang, Zilong and Zhao, Zhen and Xu, Xiaogang and Feng, Jiashi and Zhao, Hengshuang},
  journal={Advances in Neural Information Processing Systems},
  volume={37},
  pages={21875--21911},
  year={2024}
}

@inproceedings{liu2021swin,
  title={Swin transformer: Hierarchical vision transformer using shifted windows},
  author={Liu, Ze and Lin, Yutong and Cao, Yue and Hu, Han and Wei, Yixuan and Zhang, Zheng and Lin, Stephen and Guo, Baining},
  booktitle={Proceedings of the IEEE/CVF international conference on computer vision},
  pages={10012--10022},
  year={2021}
}

@article{li2024crossfuse,
  title={CrossFuse: A novel cross attention mechanism based infrared and visible image fusion approach},
  author={Li, Hui and Wu, Xiao-Jun},
  journal={Information Fusion},
  volume={103},
  pages={102147},
  year={2024},
  publisher={Elsevier}
}

@inproceedings{lin2017feature,
  title={Feature pyramid networks for object detection},
  author={Lin, Tsung-Yi and Doll{\'a}r, Piotr and Girshick, Ross and He, Kaiming and Hariharan, Bharath and Belongie, Serge},
  booktitle={Proceedings of the IEEE conference on computer vision and pattern recognition},
  pages={2117--2125},
  year={2017}
}

@article{xie2021segformer,
  title={SegFormer: Simple and efficient design for semantic segmentation with transformers},
  author={Xie, Enze and Wang, Wenhai and Yu, Zhiding and Anandkumar, Anima and Alvarez, Jose M and Luo, Ping},
  journal={Advances in neural information processing systems},
  volume={34},
  pages={12077--12090},
  year={2021}
}

@article{tong2023functional,
  title={Functional linear regression with Huber loss},
  author={Tong, Hongzhi},
  journal={Journal of Complexity},
  volume={74},
  pages={101696},
  year={2023},
  publisher={Elsevier}
}

@misc{eustaforEuropeanState,
	author = {{European State Forest Association}},
	title = {{E}uropean {S}tate {F}orest {A}ssociation - {S}taatsbetrieb {S}achsenforst --- eustafor.eu},
	howpublished = {\url{https://eustafor.eu/members/sachsenforst-state-forests-of-saxony/}},
	year = {},
	note = {[Accessed 04-12-2025]},
}

@article{hering2005conversion,
  title={Conversion of substitute tree species stands and pure spruce stands in the Ore Mountains in Saxony},
  author={Hering, S and Irrgang, S},
  journal={Journal of Forest Science},
  volume={51},
  number={11},
  pages={519--525},
  year={2005}
}

@misc{EEA_CLC2018,
  author = {{European Environment Agency (EEA)}},
  year = {2019},
  title = {{Corine Land Cover 2018 (vector), version 20}},
  publisher = {{Copernicus Land Monitoring Service}},
  howpublished = {\url{https://sdi.eea.europa.eu/catalogue/copernicus/api/records/960998c1-1870-4e82-8051-6485205ebbac?language=all}},
  note = {{Accessed on 10-11-2025}},
  doi = {10.2909/71c95a07-e296-44fc-b22b-415f42acfdf0},
  url = {https://doi.org/10.2909/960998c1-1870-4e82-8051-6485205ebbac}
}

@article{flynn2024uav,
  title={UAV-derived greenness and within-crown spatial patterning can detect ash dieback in individual trees},
  author={Flynn, WRM and Grieve, SWD and Henshaw, AJ and Owen, HJF and Buggs, RJA and Metheringham, CL and Plumb, WJ and Stocks, JJ and Lines, ER},
  journal={Ecological Solutions and Evidence},
  volume={5},
  number={2},
  pages={e12343},
  year={2024},
  publisher={Wiley Online Library}
}

@article{tan2024exploring,
  title={Exploring the Potential of GEDI in Characterizing Tree Height Composition Based on Advanced Radiative Transfer Model Simulations},
  author={Tan, Shen and Zhang, Yao and Qi, Jianbo and Su, Yanjun and Ma, Qin and Qiu, Jinghao},
  journal={Journal of Remote Sensing},
  volume={4},
  pages={0132},
  year={2024},
  publisher={AAAS}
}

@article{puletti2020lidar,
  title={Lidar-based estimates of aboveground biomass through ground, aerial, and satellite observation: a case study in a Mediterranean forest},
  author={Puletti, Nicola and Grotti, Mirko and Ferrara, Carlotta and Chianucci, Francesco},
  journal={Journal of Applied Remote Sensing},
  volume={14},
  number={4},
  pages={044501--044501},
  year={2020},
  publisher={Society of Photo-Optical Instrumentation Engineers}
}

@article{chong2022monodistill,
  title={Monodistill: Learning spatial features for monocular 3d object detection},
  author={Chong, Zhiyu and Ma, Xinzhu and Zhang, Hong and Yue, Yuxin and Li, Haojie and Wang, Zhihui and Ouyang, Wanli},
  journal={arXiv preprint arXiv:2201.10830},
  year={2022}
}

@article{chen2022bevdistill,
  title={Bevdistill: Cross-modal bev distillation for multi-view 3d object detection},
  author={Chen, Zehui and Li, Zhenyu and Zhang, Shiquan and Fang, Liangji and Jiang, Qinhong and Zhao, Feng},
  journal={arXiv preprint arXiv:2211.09386},
  year={2022}
}

@article{alidoost20192d,
  title={2D image-to-3D model: Knowledge-based 3D building reconstruction (3DBR) using single aerial images and convolutional neural networks (CNNs)},
  author={Alidoost, Fatemeh and Arefi, Hossein and Tombari, Federico},
  journal={Remote sensing},
  volume={11},
  number={19},
  pages={2219},
  year={2019},
  publisher={MDPI}
}

@article{chen2023integrating,
  title={Integrating Topographic Skeleton into Deep Learning for Terrain Reconstruction from GDEM and Google Earth Image},
  author={Chen, Kai and Wang, Chun and Lu, Mingyue and Dai, Wen and Fan, Jiaxin and Li, Mengqi and Lei, Shaohua},
  journal={Remote Sensing},
  volume={15},
  number={18},
  pages={4490},
  year={2023},
  publisher={MDPI}
}

@misc{bundDigitalTwin,
	author = {{Bundesamt f{\"u}r Kartographie und Geod{\"a}sie (BKG)}},
	title = {{B}{K}{G} - {D}igital {T}win --- bkg.bund.de},
	howpublished = {\url{https://www.bkg.bund.de/EN/Topics/Digital-Twin/digital-twin.html}},
	year = {},
	note = {[Accessed 06-12-2025]},
}

@article{Khan2025,
  doi       = {10.21105/joss.08056},
  url       = {https://doi.org/10.21105/joss.08056},
  year      = {2025},
  publisher = {The Open Journal},
  volume    = {10},
  number    = {114},
  pages     = {8056},
  author    = {Khan, Taimur and Arnold, Caroline and Grover, Harsh},
  title     = {DeepTrees: Tree Crown Segmentation and Analysis in Remote Sensing Imagery with PyTorch},
  journal   = {Journal of Open Source Software}
}

@misc{sheikholeslami2019ablation,
  title={Ablation programming for machine learning},
  author={Sheikholeslami, Sina},
  year={2019}
}

@misc{BKG_DOP_DE,
  author = {{Bundesamt f{\"{u}}r Kartographie und Geod{\"{a}}sie (BKG)}},
  title = {Digitale Orthophotos (DOP20) der Bundesrepublik Deutschland, Bodenaufl{\"{o}}sung 20 cm},
  organization = {GeoBasis-DE / BKG},
  howpublished = {\url{https://gdz.bkg.bund.de/}},
  note = {Quellenvermerk: \textcopyright\ GeoBasis-DE / BKG (JAHR\_DER\_AUFNAHME). Abruf {\"{u}}ber das Geodatenzentrum (GDZ). [Accessed 06-12-2025]},
}

@misc{sachsenLuftbildProdukteOffene,
	author = {{GeoSN}},
	title = {Luftbild-Produkte (Offene Geodaten)},
	organization = {Landesvermessung Sachsen},
	howpublished = {\url{https://www.geodaten.sachsen.de/luftbild-produkte-3995.html}},
	note = {[Accessed 2025-12-10]},
}

@misc{sachsenFachlicheDetails,
	author = {{GeoSN}},
	title = {Fachliche Details (Geobasisinformation)},
	organization = {Landesvermessung Sachsen},
	howpublished = {\url{https://www.landesvermessung.sachsen.de/fachliche-details-8645.html}},
	note = {[Accessed 2025-12-10]},
}

@article{candan2023udi,
  title={U-Net-based RGB and LiDAR image fusion for road segmentation},
  author={Candan, Arda Taha and Kalkan, Habil},
  journal={Signal, Image and Video Processing},
  volume={17},
  number={6},
  pages={2837--2843},
  year={2023},
  publisher={Springer}
}

@article{ressl2012applying,
  title={Applying 3D affine transformation and least squares matching for airborne laser scanning strips adjustment without GNSS/IMU trajectory data},
  author={Ressl, Camillo and Pfeifer, Norbert and Mandlburger, Gottfried},
  journal={The International Archives of the Photogrammetry, Remote Sensing and Spatial Information Sciences},
  volume={38},
  pages={67--72},
  year={2012},
  publisher={Copernicus GmbH}
}

@article{burns2024multi,
  title={Multi-resolution gridded maps of vegetation structure from GEDI},
  author={Burns, Patrick and Hakkenberg, Christopher R and Goetz, Scott J},
  journal={Scientific Data},
  volume={11},
  number={1},
  pages={881},
  year={2024},
  publisher={Nature Publishing Group UK London}
}

@article{xiang2025forestformer3d,
  title={ForestFormer3D: A Unified Framework for End-to-End Segmentation of Forest LiDAR 3D Point Clouds},
  author={Xiang, Binbin and Wielgosz, Maciej and Puliti, Stefano and Kr{\'a}l, Kamil and Kr{\r{u}}{\v{c}}ek, Martin and Missarov, Azim and Astrup, Rasmus},
  journal={arXiv preprint arXiv:2506.16991},
  year={2025}
}

@Article{         harris2020array,
 title         = {Array programming with {NumPy}},
 author        = {Charles R. Harris and K. Jarrod Millman and St{\'{e}}fan J.
                 van der Walt and Ralf Gommers and Pauli Virtanen and David
                 Cournapeau and Eric Wieser and Julian Taylor and Sebastian
                 Berg and Nathaniel J. Smith and Robert Kern and Matti Picus
                 and Stephan Hoyer and Marten H. van Kerkwijk and Matthew
                 Brett and Allan Haldane and Jaime Fern{\'{a}}ndez del
                 R{\'{i}}o and Mark Wiebe and Pearu Peterson and Pierre
                 G{\'{e}}rard-Marchant and Kevin Sheppard and Tyler Reddy and
                 Warren Weckesser and Hameer Abbasi and Christoph Gohlke and
                 Travis E. Oliphant},
 year          = {2020},
 month         = sep,
 journal       = {Nature},
 volume        = {585},
 number        = {7825},
 pages         = {357--362},
 doi           = {10.1038/s41586-020-2649-2},
 publisher     = {Springer Science and Business Media {LLC}},
 url           = {https://doi.org/10.1038/s41586-020-2649-2}
}

\end{document}